\documentclass[conference]{IEEEtran}
\IEEEoverridecommandlockouts

\usepackage{cite}
\usepackage{color}
\usepackage{amsmath,amssymb,amsfonts}
\usepackage{multirow}
\usepackage{graphicx}
\usepackage{textcomp}
\usepackage{xcolor}
\usepackage{hyperref}

\usepackage[linesnumbered, ruled]{algorithm2e}
\usepackage{setspace}
\usepackage{algpseudocode}
\algnewcommand{\LeftComment}[1]{\Statex \(\triangleright\) #1}

\usepackage{amsmath}

\def\BibTeX{{\rm B\kern-.05em{\sc i\kern-.025em b}\kern-.08em
    T\kern-.1667em\lower.7ex\hbox{E}\kern-.125emX}}
\begin{document}
\title{Improving Communication Efficiency of Federated Distillation via Accumulating Local Updates
}
\author{
	Zhiyuan Wu$^{1,2}$, Sheng Sun$^{1}$, Yuwei Wang$^1$\thanks{\textit{(Corresponding author: Yuwei Wang)}}, Min Liu$^{1,3}$, Tian Wen$^{1}$, Wen Wang$^{1,2}$ and Bo Gao$^{4}$\\
	$^1$Institute of Computing Technology, Chinese Academy of Sciences \\
	$^2$University of Chinese Academy of Sciences \quad $^3$Zhongguancun Laboratory \quad $^4$Beijing Jiaotong University \\
	\{wuzhiyuan22s,sunsheng,ywwang,liumin\}@ict.ac.cn \\ marrowd611@gmail.com \quad wangwen22s@ict.ac.cn \quad bogao@bjtu.edu.cn\\
}

\maketitle

\begin{abstract}
As an emerging federated learning paradigm, federated distillation enables communication-efficient model training by transmitting only small-scale knowledge during the learning process. 
To further improve the communication efficiency of federated distillation, we propose a novel technique, ALU, which accumulates multiple rounds of local updates before transferring the knowledge to the central server. ALU drastically decreases the frequency of communication in federated distillation, thereby significantly reducing the communication overhead during the training process. Empirical experiments demonstrate the substantial effect of ALU in improving the communication efficiency of federated distillation.
\end{abstract}

\begin{IEEEkeywords}
Federated Learning, Knowledge Distillation, Efficient Communication
\end{IEEEkeywords}

\section{Introduction}
With the growing demand for intelligent services and the increasing awareness of data privacy, federated learning has emerged as a crucial approach for conducting distributed machine learning without sharing the raw data of participants. A major challenge of prevalent federated learning methodologies is their communication efficiency \cite{mcmahan2017communication,kairouz2021advances}, as they require participants, referred to as clients, to periodically transmit extensive model parameters to a central server. This becomes particularly problematic in scenarios with bandwidth limitations or high communication costs. To address this issue, federated distillation \cite{wu2023fedict,jeong2018federated} is proposed as a communication-efficient federated learning paradigm, which only transfers compact knowledge extracted from clients during the training process, circumventing the need for large-scale parameters transmission. Despite the great success of federated distillation, potentials remain to enhance communication efficiency by decreasing the frequency of knowledge exchange.

In this paper, we propose a novel technique, ALU, to improve the communication efficiency of federated distillation without compromising model accuracy. The essence of ALU is to delay the transmission of knowledge in federated distillation until multiple rounds of local model updates are accumulated on the client side, after which a single round of more informative knowledge is transmitted and received to and from the server.
To our best knowledge, \textbf{this paper is the first work to improve the communication efficiency of federated distillation from the perspective of reducing communication frequency.} The proposed ALU allows for flexible integration with prevailing state-of-the-art federated distillation methods, while still preserving user privacy, accommodating model heterogeneity, and upholding other inherent benefits of federated distillation.

\begin{figure}[b]
	\centering
	\includegraphics[width=0.5\textwidth]{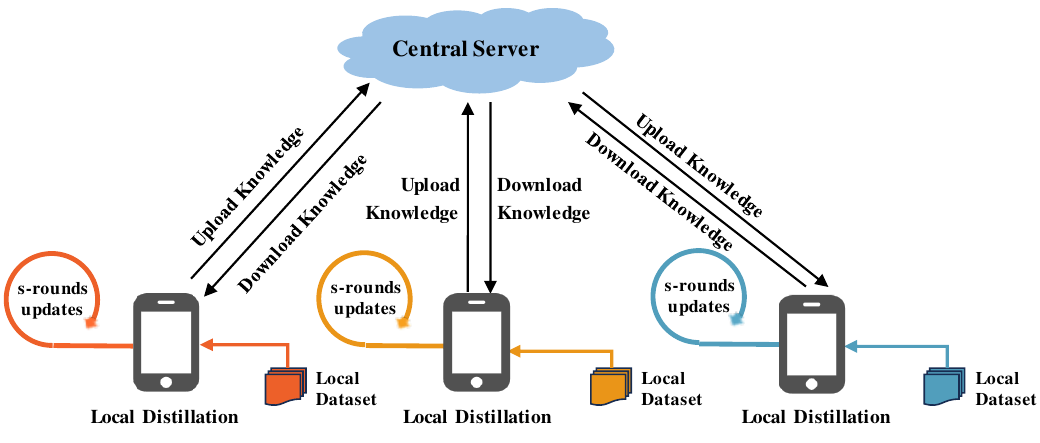}
	\caption{Illustration of ALU in federated distillation with three clients.}
	\label{main-figure}
\end{figure}

\section{Proposed Method}
During the $t$-th communication round in federated distillation, client $k$ optimizes its local model parameters $W^k$ with a linear combination of the cross-entropy loss $L_{CE}(\cdot)$ on its local data $\mathcal{D}^k$ and the distillation loss $L_{KD}(\cdot)$ over the global knowledge downloaded from the server, that is:
\begin{equation}
    \mathop {\min }\limits_{{W^k}} \mathop E\limits_{({X_i^k},{y_i^k})\sim{\mathcal{D}^k}} [J_{CE}+\beta \cdot J_{KD}],
\end{equation}
subject to 
\begin{equation}
\left\{ \begin{array}{l}
{J_{CE}} = {L_{CE}}(f({W^k};X_i^k),y_i^k))\\
{J_{KD}} = {L_{KD}}(f({W^k};X_i^k),{R_t}(X_i^k))
\end{array} \right.,
\end{equation}
where $X_i^k$, $y^k_i$ denotes the $i$-th sample and label in $\mathcal{D}^k$, $f(W^*;\cdot)$ denotes the nonlinear function of a neural network determined by model parameters $W^*$, $\beta$ is the distillation weighting factor, and $R_t(\cdot)$ is the download global knowledge in round $t$ associated with $X_i^k$. The global knowledge can be averaged logits from other clients \cite{jeong2018federated}, the output of the server model \cite{wu2023fedict,wu2022exploring}, or fetched from remote knowledge cache \cite{wu2023fedcache}.

In this paper, we accumulate local updates in federated distillation by $s$ rounds (formulating s-ALU), which is illustrated in Fig. \ref{main-figure}. When the number of communication rounds cannot be evenly divided by $s$, the latest downloaded knowledge is adopted for distillation-based optimization, and the knowledge uploading as well as downloading is delayed until the current round number reaches an integer multiple of $s$, that is:
\begin{equation}
    {J_{{ALU} - KD}} = {L_{KD}}(f({W^k};X_i^k),{R^{ALU}_t}(X_i^k)),
\end{equation}
where
\begin{equation}
    {R^{ALU}_t}(X_i^k) = {R_{t - t\bmod s}}(X_i^k).
\end{equation}
With the integration of our proposed ALU, federated distillation forms a new optimization objective on clients dispense with communication in every round, that is:
\begin{equation}
    \mathop {\min }\limits_{{W^k}} \mathop E\limits_{({X_i^k},{y_i^k})\sim{\mathcal{D}^k}} [J_{CE}+\beta \cdot J_{ALU-KD}].
\end{equation}
Note that when $s=1$, ALU degraded to no additional operation.

\begin{figure}[t]
	\centering
	\includegraphics[width=0.5\textwidth]{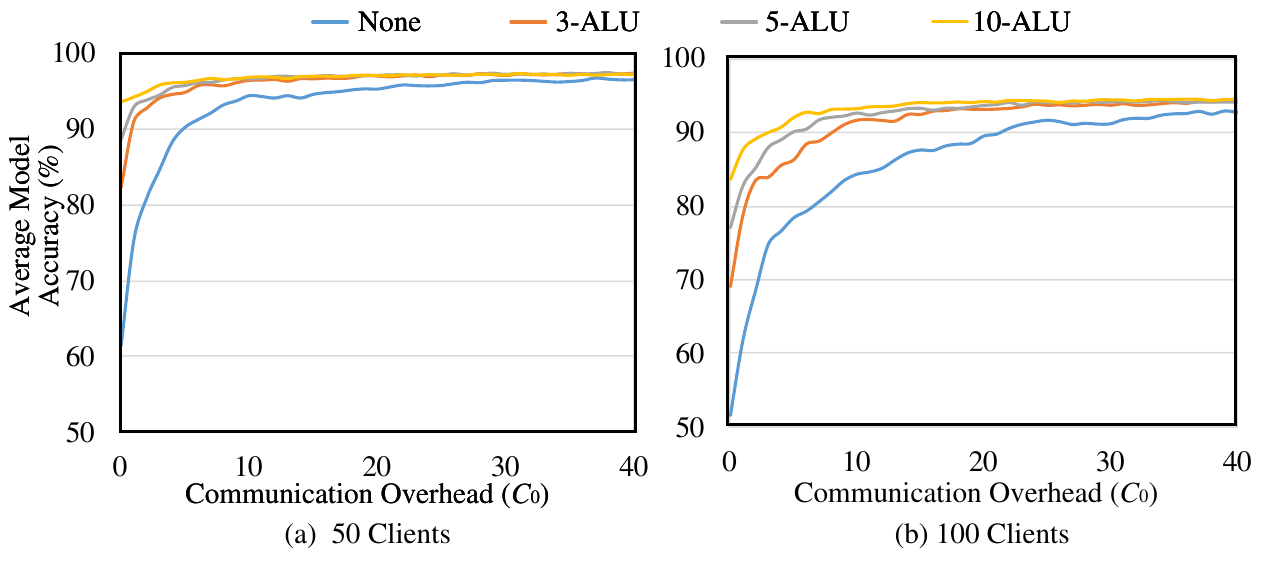}
	\caption{Comparison of average model accuracy (\%) per unit of communication with different settings. The technique corresponding to the label names is incorporated with FedCache to obtain the corresponding learning curve. The same as below.}
	\label{communication}
\end{figure}

\section{Preliminary Results}
We conduct simulation experiments on the MNIST \cite{lecun1998gradient} dataset, incorporating our proposed ALU into FedCache \cite{wu2023fedcache} with either 50 or 100 clients. Our hyper-parameters maintain the same as those employed in the main experiment of literature \cite{wu2023fedcache}. In addition, we set $s$ to be a member of $\{3,5,10\}$ and record both the average model accuracy after convergence and the average model accuracy per unit of communication.
To provide a more intuitive understanding, we exclusively focus on the communication that occurred during the training process. This communication is quantified in units of $C_{0}$, which represents a single-round training communication burden of FedCache.

Fig. \ref{communication} illustrates the effect of ALU on average model accuracy per unit of communication. Our experimental comparisons reveal steeper learning curves for FedCache integrated with ALU. In addition, all curves converged to similar final values. These phenomena demonstrate that ALU is instrumental in reducing communication overhead without compromising model accuracy. Essentially, ALU enhances the efficiency of transmitted information per unit of communication by enabling more informative and sufficient knowledge exchange through accumulating local updates with less frequent communication.

\begin{figure}[t]
	\centering
	\includegraphics[width=0.38\textwidth]{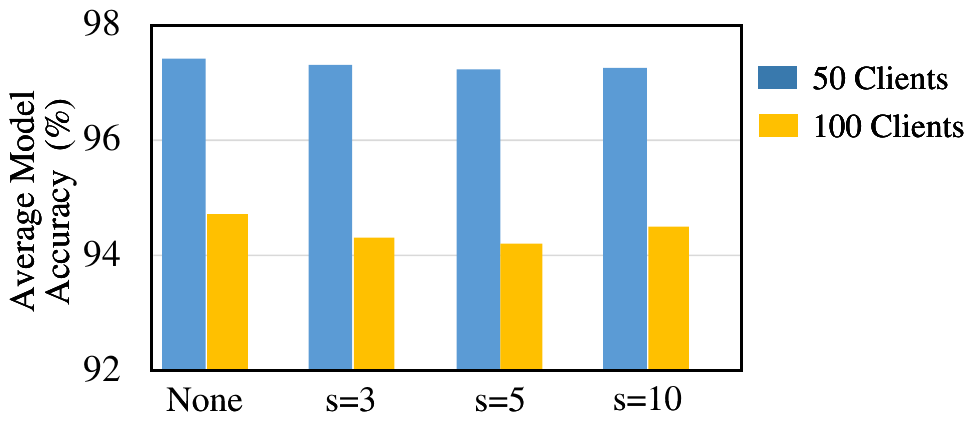}
	\caption{Comparison of average model accuracy (\%) after convergence with different settings.}
	\label{accuracy}
\end{figure}

Moreover, Fig. \ref{accuracy} displays the influence of ALU on converged average model accuracy. This figure reveals that integrating ALU into federated distillation methods results in a negligible decline in average accuracy upon model convergence. This finding confirms ALU's significant effect in the judicious use of communication resources on the premise of preserving the integrity of model performance.

\section{Conclusion}
This paper proposes the ALU technique to improve the communication efficiency of federated distillation. By delaying knowledge exchange until accumulating multiple local updates, ALU effectively de-frequency the communication between clients and the server, thereby reducing the bandwidth requirements that are critical in communication-constrained scenarios. Experiments demonstrate that ALU can be easily incorporated with existing federated distillation methods, and can improve communication efficiency while preserving model accuracy.

\bibliographystyle{IEEEtran}

\vspace{12pt}

\end{document}